\def\year{2023}\relax
\documentclass[letterpaper]{article} 
\usepackage{aaai23}  
\usepackage{times}  
\usepackage{helvet} 
\usepackage{courier}  
\usepackage[hyphens]{url}  
\usepackage{graphicx} 
\usepackage{natbib}  
\usepackage{caption} 
\usepackage{adjustbox}
\usepackage{setspace}
\usepackage{amssymb}
\usepackage{comment}
\usepackage{amsmath,amssymb,cases} 
\usepackage{color}

\usepackage{subfigure}
\usepackage{booktabs} 
\usepackage{multirow}
\usepackage{multicol}
 \usepackage{enumitem}
\usepackage{algorithm}  
\usepackage[noend]{algpseudocode}

\usepackage{newfloat}
\usepackage{listings}
\DeclareCaptionStyle{ruled}{labelfont=normalfont,labelsep=colon,strut=off} 
\floatstyle{ruled}
\newfloat{listing}{tb}{lst}{}
\floatname{listing}{Listing}

\usepackage{lipsum}

\newcommand\blfootnote[1]{%
  \begingroup
  \renewcommand\thefootnote{}\footnote{#1}%
  \addtocounter{footnote}{-1}%
  \endgroup
}

\usepackage{bbold}
\usepackage{siunitx}    

\usepackage{etoolbox}
\robustify\bfseries

\newcommand{\eqnref}[1]{Eq.~(\ref{#1})}

\newcommand{\tableref}[1]{Table~\ref{#1}} 

\newcommand{\figref}[1]{Figure~\ref{#1}} 

\newtheorem{definition}{Definition}
\graphicspath{{image/}}
 \usepackage{enumitem}
\usepackage{fancyhdr}
\title{Easy Begun is Half Done: \\ Spatial-Temporal Graph Modeling with ST-Curriculum Dropout}

\author{
    \parbox{\linewidth}{\centering
        Hongjun Wang\textsuperscript{\rm 1,2\textsection},
        Jiyuan Chen\textsuperscript{\rm 1,2\textsection},
        Tong Pan\textsuperscript{\rm 1,2,3},	
        Zipei Fan\textsuperscript{\rm 4 $\dagger$}, 	
        Boyuan Zhang\textsuperscript{\rm 1,2}
        Renhe Jiang\textsuperscript{\rm 4,5},
        Lingyu Zhang\textsuperscript{\rm 1,2}, 	
        Yi Xie\textsuperscript{\rm 6}, 		
        Zhongyi Wang\textsuperscript{\rm 6}, 		
        and Xuan Song\textsuperscript{\rm 1,2$\dagger$}, 	
    } \\
    {}
}
\affiliations{
    \textsuperscript{\rm 1} SUSTech-UTokyo Joint Research Center on Super Smart City, Department of Computer Science and Engineering, Southern University of Science and Technology;\\
    \textsuperscript{\rm 2} Research Institute of Trustworthy Autonomous Systems,  SUSTech;\\
    \textsuperscript{\rm 3} Department of Physics, The Chinese University of Hong Kong;
     \textsuperscript{\rm 4} Center for Spatial Information Science, University of Tokyo; \textsuperscript{\rm 5} Information Technology Center, University of Tokyo \\
     \textsuperscript{\rm 6} Huawei Technologies CO.LTD;\\
 
}

\usepackage{bibentry}
\begin{document}

\maketitle
\begingroup\renewcommand\thefootnote{\textsection }
\footnotetext{Equal contribution.}
\endgroup
\begingroup\renewcommand\thefootnote{$\dagger$}
\footnotetext{Corresponding to: Zipei Fan (fanzipei@iis.u-tokyo.ac.jp) and Xuan Song (songx@sustech.edu.cn).}
\endgroup

\thispagestyle{fancy} 
\lhead{} 
\chead{} 
\rhead{} 
\lfoot{} 
\cfoot{} 
\rfoot{\thepage} 
\renewcommand{\headrulewidth}{0pt} 
\renewcommand{\footrulewidth}{0pt} 
\pagestyle{fancy}
\rfoot{\thepage}
 \begin{abstract}
 
Spatial-temporal (ST) graph modeling, such as traffic speed forecasting and taxi demand prediction, is an important task in deep learning area.
However, for the nodes in graph, their ST patterns can vary greatly in difficulties for modeling, owning to the heterogeneous nature of ST data. We argue that unveiling the nodes to the model in a meaningful order, from easy to complex, can provide performance improvements over traditional training procedure. The idea has its root in Curriculum Learning \cite{bengio2009curriculum} which suggests in the early stage of training models can be sensitive to noise and difficult samples. In this paper, we propose ST-Curriculum Dropout, a novel and easy-to-implement strategy for spatial-temporal graph modeling. Specifically, we evaluate the learning difficulty of each node in high-level feature space and drop those difficult ones out to ensure the model only needs to handle fundamental ST relations at the beginning, before gradually moving to hard ones. Our strategy can be applied to any canonical deep learning architecture without extra trainable parameters, and extensive experiments on a wide range of datasets are conducted to illustrate that, by controlling the difficulty level of ST relations as the training progresses, the model is able to capture better representation of the data and thus yields better generalization. 

 \end{abstract}
 
 \section{INTRODUCTION}

\begin{figure}[t]
	\centering
	\includegraphics[width=1\linewidth]{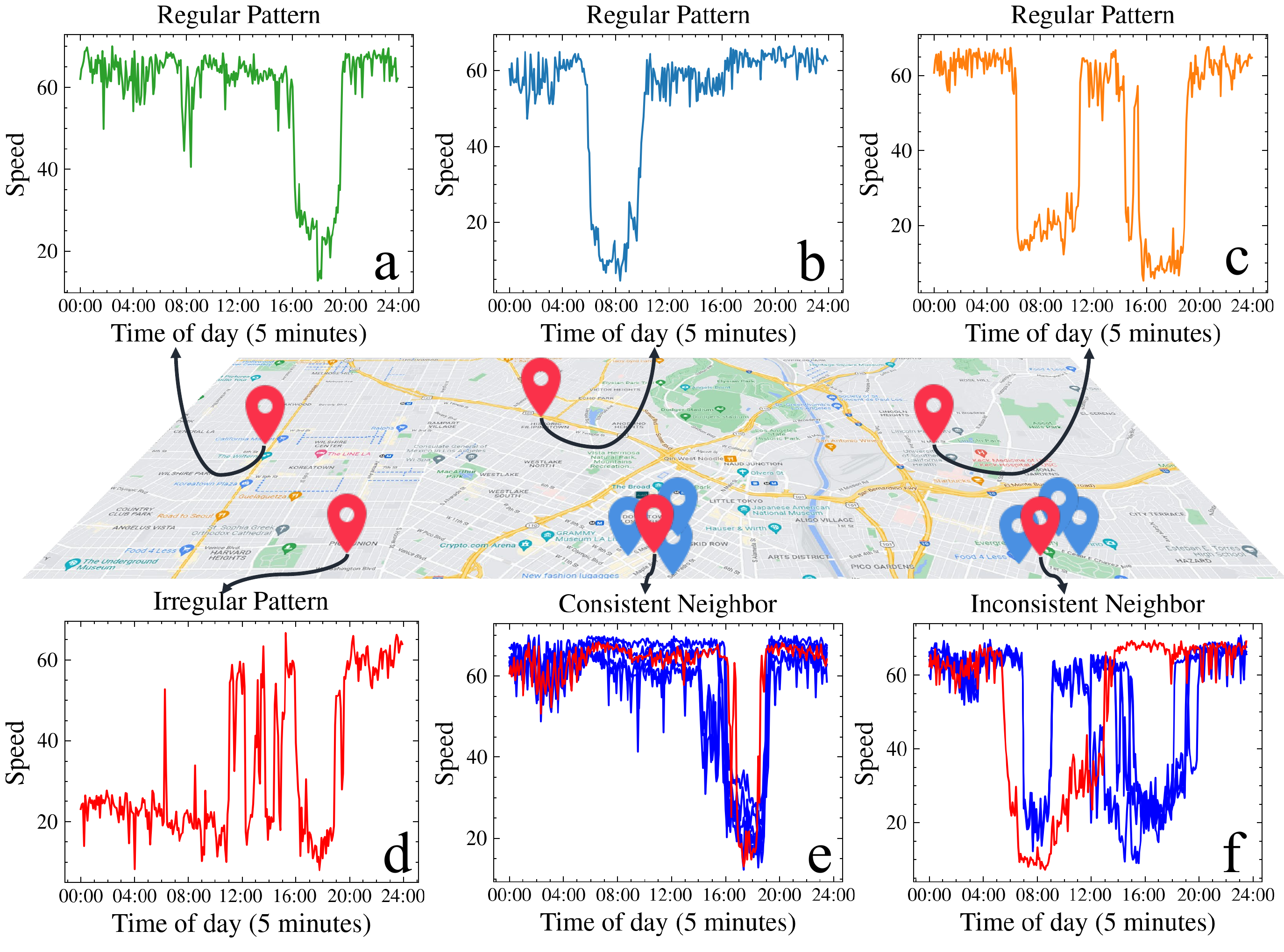}
	\caption{Speed data visualization in METR-LA dataset. }\label{fig:motivation}
\end{figure}

With the advances in deep learning techniques such as Convolutional Neural Network (CNN) and Graph Neural Networks (GNN), spatial-temporal graph modeling has been receiving increasing attention. The basic assumption behind it is that a node’s future state is conditioned on its historical state as well as its spatial neighbors’ historical state \cite{wu2019graph}. Thus, the modeling target seeks to discover both the node-level spatial inter-dependent relations and the node's self-dependent temporal relations. The task has many  real-world applications like traffic speed forecasting \cite{yu2017spatio} and human action recognition \cite{yan2018spatial}.


While plenty of attention has been devoted to designing more sophisticated model architectures so as to capture better ST relations, we are convinced that it's also valuable to take a step back and focus on an essential yet easily-overlooked question: \textit{Can we simply use a more advanced training strategy to improve model performances? } Almost all the existing work in ST modeling tends to treat the nodes in the graph equally and feed them into the neural network. However, chances are that the difficulty of modeling a node's ST pattern can vary greatly. \figref{fig:motivation} (a), (b) and (c) show the regular speed pattern of nodes in METR-LA dataset \cite{li2017diffusion}, which demonstrate that most of the nodes are only significantly affected by the morning and/or evening peaks, when heavy traffic congestions might happen. These regular temporal patterns are easy for models to learn. Nevertheless, \figref{fig:motivation} (d) shows an example of nodes whose speed patterns are highly disordered and unpredictable, and their data representation can be very different from the majority of nodes. This is an irregular one which is hard to model. For \figref{fig:motivation} (e) and (f), as we mentioned before, ST graph modeling also relies on the node-level spatial inter dependency. \figref{fig:motivation} (e) shows the circumstance where a node (red line) and its first-order spatial neighbors' (blue lines) speed patterns are very alike. This spatial relation without doubt can be easily captured by the model, while \figref{fig:motivation} (f) shows the opposite way where the node (red line) doesn't share the same trend as its neighbors (blue lines). Although in \figref{fig:motivation} (f) the temporal speed patterns of both the node and its neighbors are easy to learn, their spatial relations can be confusing for the model especially at early training stages. Extensive research has suggested that presenting the training samples in a meaningful order, from easy to hard, can benefit the training process and such strategy is referred to as Curriculum Learning \cite{kumar2010self,xu2020curriculum, wang2021curgraph}.

Curriculum Learning (CL) follows the intuitions of humans: learning should start with the fundamental knowledge and then gradually build on that to handle more complicated ones. The idea is first proposed by  \cite{bengio2009curriculum} and we've witnessed its successful applications in many fields like image classification \cite{srivastava2014dropout}, text classification \cite{ma2017self} and object segmentation \cite{kumar2011learning}. From the view of optimization, curriculum learning excludes the impacts from difficult or even noisy samples at the early stage of training when the model is vulnerable, and is thus able to prevent the model
from getting stuck in bad local minima, making it converge faster and reach a better solution than the standard approach. The success of curriculum learning, especially in highly non-convex deep models, gives us the motivation to introduce it into the task of spatial-temporal graph modeling, and to the best of our knowledge, such an idea still remains unexplored.

In previous work, the smallest unit to be assigned a difficulty level is a data sample (e.g., for image classification, a data sample should be a single image), and the curriculum is built by progressively increasing the difficulty threshold for input data. However, in the context of spatial-temporal graph modeling, a data sample is a graph consisting of multiple spatially-connected nodes with their temporal features. Therefore, it is hard to rank two data samples' difficulties when one's nodes suffer from messy spatial relations, and the others' nodes suffer from terribly orderless temporal patterns. In a word, as a graph, the data sample here lacks the necessary atomicity to be analyzed as a whole. On the contrary, we find that the node in graph is atomic and itself contains sufficient ST information (spatial neighborhood relationships in spatial view and its own time series values in temporal view). Therefore, unlike previous work, we innovatively set the smallest unit as a node instead of the whole graph and define our curriculum as gradually introducing the nodes with complex ST relations to the model.

The key challenge towards utilizing curriculum learning lies in determining easy/hard nodes. In this paper, we propose to evaluate the difficulty of nodes in high dimensional feature space. Specifically, we define easy nodes as those whose data representations locate in high-density area of the feature space and are consistent with their spatial neighbors on graph (we denote the neighbors on graph as \textbf{$\textbf{G}$-neighbors} for simplicity). Conversely, difficult nodes have their data representation either solitarily far away from the majority or inconsistent with its $G$-neighbors. Then based on these difficulty scores, we propose our ST-Curriculum Dropout as a novel training strategy specially for spatial-temporal graph modeling. The idea is inspired by \cite{sinha2020curriculum} which smooths the feature space with a continuously weakened Gaussian kernel. We take a braver step by directly dropping out the nodes with sophisticated ST relations at the early stage of training, such that the model can only propagate and learn the most fundamental ST relations. We further propose a smooth curriculum arrangement function to progressively introduce harder nodes into the model, allowing at each training step, the model can focus on ``interesting" ST relations, that are near its border of capability, neither too easy nor too hard, to expand the border gradually.  

\noindent The main contributions in this paper are listed as follows:
\begin{itemize}[leftmargin=*]
	\item[$\bullet$] We explore and make novel adaptions of CL into spatial-temporal graph modeling. To the best of our knowledge, this is the first time curriculum learning is proven to be prospective in such a task.  
	\item[$\bullet$] We propose a novel CL training strategy, which can also be formulated as an external plug-in for canonical deep learning models, along with a smooth CL arrangement function for ST graph modeling task.
	\item[$\bullet$]
	We obtained universal performance gain on a wide range of real-world datasets, illustrating the effectiveness of our design. 
\end{itemize}

\section{Preliminaries}\label{sec:prelim}


\begin{figure*}[t]
	\centering
	\includegraphics[width=1\linewidth]{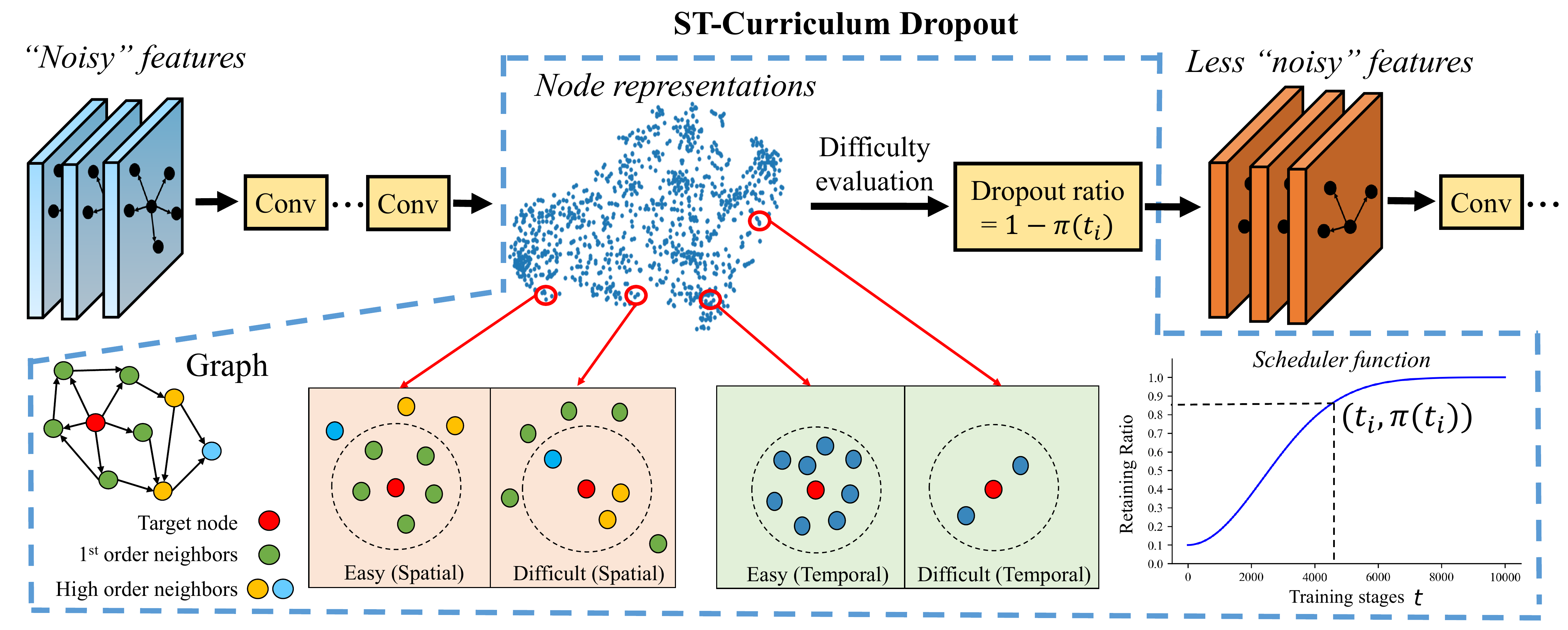}
	\caption{The schematic of ST-Curriculum Dropout is shown above as a plug-in to deep learning models. In high-level feature space, nodes with difficult ST-relations will be dropped out at a certain ratio to “denoise” the data, ensuring a smoother objective for the model to optimize at early training stages. The ratio to retain difficult nodes will be progressively increased during training until the overall utilization.
	 }\label{fig:framework}
\end{figure*}

\begin{definition}
\textbf{Graph}. In this paper, a graph is represented as $\mathcal{G}=(\mathcal{V},\mathcal{E}, A)$, where  $\mathcal{V}$ denotes the set of nodes, $\mathcal{E} \subseteq \mathcal{V}\times \mathcal{V} $ indicates the set of edges, and $A$ is the adjacency matrix derived from graph $\mathcal{G}$. At each time step $t$, the graph $\mathcal{G}$ has a dynamic feature matrix $\mathbf{X}_t \in \mathbb{R}^{|\mathcal{V}| \times \mathcal{D}}$ and $\mathcal{D}$ is the dimension of node features. \blfootnote{The related work of this paper is included in Appendix A.1}
\end{definition}

\begin{definition}
\textbf{Spatial-Temporal Graph Modeling}. Given a graph $\mathcal{G}$ and its N step historical observation $\mathbf{X}_{t-N: t}$, the task of ST graph modeling is to 
learn a neural network $f$ which is able to predict the future $S$ step graph features $\overline{\mathbf{X}}_{t+1: t+S}$. We  summarize the mapping relation as: 
$$f: \left[\mathbf{X}_{(t-N): t}, \mathcal{G}\right] \mapsto \overline{\mathbf{X}}_{(t+1):(t+S)}.$$
\end{definition}

%
%
%
%

 \section{METHODOLOGY}
 In this section, we decompose our ST-Curriculum Dropout (abbreviated as STC-Dropout) into two stages: Difficulty Evaluation and Curriculum Learning. In the first stage, each node in graph will be assigned a score reflecting its difficulty of ST relations. This is done by measuring both a node's data distribution density (temporal view) and its consistency with $G$-neighbors (spatial view). In the second stage, based on these scores, nodes are exposed to the model in an easy-to-difficult fashion, resulting in the final curriculum that the model will be trained on. The schematic of our method is shown in \figref{fig:framework} as a plug-in and the details will be elaborated next.
        


\subsection{Difficulty Evaluation}\label{sec:diffeval}
At the core of our idea is to develop a general difficulty evaluation justified for all possible models in spatial-temporal graph modeling tasks, even though in this paper only GNNs are conducted as an example. We analyze the nodes' high-level representations retrieved from previous layers in the model. Here, one might question why we don't directly analyze the original data itself rather than its high-level representation. However, this is non-trivial since models always encode the data through advanced non-linear transformations, so what might seem reasonable to a human in the original data might not be considered the same way by the model in its high-dimension feature space. Therefore, we argue that the difficulty scores should be derived from the high-level data representation inside the model so that we can ``eliminate'' what the model itself thinks difficult or disturbing to its training. This idea shares a consensus with Self-Paced-Learning \cite{kumar2010self}, where the author claims the models should take charge in deciding the difficulties of samples. 

With the node's high-level representation which infers the specific semantic information perceived by the model, we can now evaluate the difficulties. For the sake of clarity, let's first remind the basic assumption of ST graph modeling: a node’s future state is conditioned on its historical state as well as its $G$-neighbors’ historical state. Intuitively, the easier nodes should be consistent with $G$-neighbors (spatial view) and have their high-level data representations in the high density area of the distribution (temporal view). On the contrary, more difficult nodes should be the ones either have data representation solitarily far away from the majority or are inconsistent with its $G$-neighbors.

Formally, let  $h^{(l)} \in \mathbb{R}^{|\mathcal{V}| \times d^{(l)}}$ indicate the $d^{(l)}$-dimensional data representation of all nodes, which is generated by the $(l-1)^{th}$ layer $f^{(l-1)}$ with $h^{(l-1)}$ as
\begin{align}
	h^{(l)}=f^{(l-1)}(h^{(l-1)}).
\end{align}
For node $v_i$ and its representation $h^{(l)}_{i} \in \mathbb{R}^{ d^{(l)}}$, we build a $d^{(l)}$-dimensional ball with
radius $R$ centered at $h^{(l)}_{i}$:
\begin{align}
	\mathcal{B}\left(h^{(l)}_{i}, R\right)=\left\{h^{(l)}_{j} \mid D(h^{(l)}_{j}, ~h^{(l)}_{i})\leq R\right\},
\end{align}
where $D$ is the distance function. We normalize the vectors here to eliminate the effect of variances along different directions and apply Cosine Distance to alleviate the curse of dimensionality in high-level feature space. Then, in order to measure the node's consistency with $G$-neighbors, we here further introduce the concept of $k$ order neighbors.
\begin{definition}
	\textbf{$k$ order neighbors}. If  node $v_j$ is a $k$  order neighbor of  node $v_i$ on graph $\mathcal{G}$, there exists a path $P_{v_i \rightarrow v_j}: v_{i} \rightarrow v_{m} \rightarrow \cdots \rightarrow v_j$ which satisfies $|P_{v_i\rightarrow v_j}|-1 \leq k$. 
\end{definition}

Let $\Omega_{v_i}^k=\{v_j \mid |P_{v_i \rightarrow v_j}|-1\leq k \}$ 
denotes the set of nodes being the $k$ order neighbors of $v_i$,  an indicator function $\mathcal{I}$ can be further defined as 
$$ \mathcal{I}_{v_i}(v_j,\Omega_{v_i}^k)=\left\{
\begin{aligned}
	1, & \qquad v_j \in \mathcal{B}\left(h^{(l)}_{i}, R\right) \wedge v_j \in \Omega_{v_i}^k \\
	0, & \qquad \text{otherwise} 
\end{aligned}
\right.
$$
We can now count the portion of node $v_i$'s $k$ order neighbors which are wrapped by the ball $\mathcal{B}\left(h^{(l)}_{i}, R\right)$ as:
\begin{align}\label{equ:spatial}
	\frac{1}{\mid\Omega_{v_i}^k\mid}\sum_{v_j \in \mathcal{V}} \mathcal{I}_{v_i}(v_j,\Omega_{v_i}^k).
\end{align}
Intuitively, the higher the fraction of node $v_i$'s $k$ order neighbors captured by the ball, the more consistent $v_i$ should be with its $G$-neighbors, which indicates an easier spatial relation for the model to capture. 
However, for different nodes, their data distribution density (i.e., if a node's high-level data representation are 
intensively surrounded by many of others) can vary by orders of magnitudes, which the simple ratio in Formula (\ref{equ:spatial}) can not properly capture. Take node $v_i$ as an example, if ball $\mathcal{B}\left(h^{(l)}_{i}, R\right)$ wraps massive nodes inside, which means $h^{(l)}_i$ is located in the high density area of the feature space, then $v_i$ is a normal node with easier temporal patterns and has many peers similar to itself. If the nodes wrapped by the ball further 
account for a large portion of $v_i$'s $k$ order neighbor, then the pattern of $v_i$ should be easy to learn in both spatial and temporal aspects. Conversely, if $v_i$'s high-level representation $h_i^{l}$ stays solitarily away from others (i.e., in low density area), then $v_i$ is an outlier in the dataset. In this case, $v_i$ should be more difficult to learn in temporal aspect regardless of its consistency with $G$-neighbors.

Therefore, in addition to Formula (\ref{equ:spatial}) which serves as a difficulty measurer in spatial aspect, we further propose Formula (\ref{equ:temporal}) to measure the node's difficulty from temporal aspect as follows: 
\begin{align}\label{equ:temporal}
    \frac{|\mathcal{B}\left(h^{(l)}_{i}, R\right)|}{|\mathcal{B}\left(h^{(l)}_{i}, R\right)|+\epsilon_{R}},
\end{align}
where $\epsilon_{R} = \frac{1}{|\mathcal{V}|}\sum_{i=1}^{|\mathcal{V}|} |\mathcal{B}\left(h^{(l)}_{i}, R\right)|$ acts as a penalty term for those ``lonely" nodes. Specifically, the functionality of Formula (\ref{equ:temporal}) mainly relies on three important properties: 1) The value of Formula (\ref{equ:temporal}) increases monotonically as the node's distribution density becomes larger (i.e., $|\mathcal{B}\left(h^{(l)}_{i}, R\right)|$ gets larger). 2) The value of $\epsilon_{R}$ represents the average data distribution density and serves as a penalizer. When $|\mathcal{B}\left(h^{(l)}_{i}, R\right)| \gg \epsilon_{R} $, we have $    \frac{|\mathcal{B}\left(h^{(l)}_{i}, R\right)|}{|\mathcal{B}\left(h^{(l)}_{i}, R\right)|+\epsilon_{R}} \approx 1$. Conversely, when $|\mathcal{B}\left(h^{(l)}_{i}, R\right)| \ll \epsilon_{R} $, we have $    \frac{|\mathcal{B}\left(h^{(l)}_{i}, R\right)|}{|\mathcal{B}\left(h^{(l)}_{i}, R\right)|+\epsilon_{R}} \approx 0$. The penalty strength of $\epsilon_{R}$ actually grows as the node's data distribution density decreases. 3) The range of Formula (\ref{equ:temporal}) is constrained 
to $[0,1]$, which is consistent with the 
range of Formula (\ref{equ:spatial}). Therefore, by integrating Formula (\ref{equ:spatial}) and Formula (\ref{equ:temporal}), we
propose our difficulty evaluation function $Diff(v_i)$ as:
\begin{small}
\begin{align}\label{equ:difficult}
	2-\underbrace{\frac{1}{\mid\Omega_{v_i}^k\mid } \sum_{v_j \in \mathcal{V}} \mathcal{I}_{v_i}(v_j,\Omega_{v_i}^k)}_{\text{Spatial difficulty}}-\underbrace{\frac{|\mathcal{B}\left(h^{(l)}_{i}, R\right)|}{|\mathcal{B}\left(h^{(l)}_{i}, R\right)|+\epsilon_{R}}}_{\text{Temporal difficulty}}.
\end{align}
\end{small}
Moreover, to avoid excessive hyperparameters, we here heuristically define the radius $R$ in \eqnref{equ:difficult} as $R=\frac{1}{|\mathcal{V}|}\sum_{i=1}^{|\mathcal{V}|} \mathcal{Q}_{\rho}(\{ D(h^{(l)}_{i}, ~h_j^{(l)}): j \in \{1, \cdots, |\mathcal{V}|\} \})$, where  $\mathcal{Q}$ is a quantile function that return the $\rho\%$ quantile values expanding from small to large.

\subsection{Curriculum Learning}

In this section, we start from the formalization of curriculum learning  \cite{bengio2009curriculum}. Let $\lambda\in[0,1]$ denotes the training time such that the  training begins at $\lambda=0$ and finishes at $\lambda=1$. Suppose $Q_\lambda(x)$ is the training distribution at time $\lambda$, where the training sample $x$ is drawn. The concept of curriculum learning requires that the training samples from $Q_\lambda$ should be easier than the samples from $Q_{\lambda+\delta}$, $\delta>0$. Mathematically, curriculum learning holds the assumption that 
\begin{align}
	Q_{\lambda}(x) \propto W_{\lambda}(x) P(x),
\end{align}
where $P(x)$ denotes the target training distribution including both easy and hard samples, and $0\leq W_{\lambda}(x)\leq 1$ is the difficult criterion for every training time $\lambda$ and sample $x$. In the meanwhile, the choice of $W_{\lambda}(x)$ should satisfy such property:
\begin{align}\label{equ:Shannon}
	H(Q_{\lambda}(x))<H(Q_{\lambda+\delta}(x)),
\end{align}
where $H$ is the Shannon’s entropy. \eqnref{equ:Shannon} expounds the fact that the diversity and information in the training set should gradually increase with respect to the training time $\lambda$, which is in accordance with our idea to progressively exposing nodes with more complicated ST relations to the model.

To control the change of difficulties of the training set during training (the pace of curricula), we follow the theory in \cite{bengio2009curriculum} and introduces our curriculum scheduler $\pi(t)$,  which indicates the retain rate of nodes at a certain training stage. Let $t \in \{0,1,2, \dots\}$ denotes the training stages, measured by gradient updates, the scheduler should smoothly level up the difficulty of training corresponding to $t$ by gradually increasing the retain rate of nodes in a given layer. More straightforwardly, we want the model to start from a small fraction of easy nodes ($\pi(0)=\bar{\alpha}$) and gradually increases the availability of nodes to complicate the task until eventually recovering to normal training ($\lim_{t\rightarrow \infty} \pi(t)=1$). The scheduler function $\pi(t)$ can be formulated as 
\begin{align}
	\pi(t)= 1 - (1- \bar{\alpha}) \exp (-\beta t),  \quad \beta>0 \label{eqn:curriculum_function}
\end{align}
With the previous derivations, we can now formally introduce our ST-Curriculum Dropout strategy as, given a training stage $t$ and nodes' high-level representation $h^l$, we keep the top $\pi(t)\%$ easiest nodes and dropout (set to zero) the others to control the level of difficulty of the task. The procedure can be mathematically written as:

\begin{align}
    S_{t}&=\mathcal{Q}_{\pi(t)}(\left\{ Diff(v_{j}  ) : v_j \in \mathcal{V}\right\})\\
	\mathcal{M}&=\mathbb{1}\left\{ Diff(v_i) \leq S_t  \mid v_i \in \mathcal{V} \right\}\label{equ:stc_drop}\\
	\widetilde{h}^{(l)}&=\mathcal{M}* h^{(l)}.\label{equ:dropout}
\end{align}
Some might argue that the choice of $\beta$ in \eqnref{eqn:curriculum_function}
can be annoying since it requires cross-validation.
However, as 
suggested in \cite{morerio2017curriculum}, this value can actually be heuristically decided. In order to reserve all the nodes to recover normal training in later stages, 
we should have $\pi(T)> 1-\frac{1}{|\mathcal{V}|}$, where $T$ refers to the total number of gradient updates. 
Surprisingly, we observe that $\beta=\frac{10^{3}}{T\times|\mathcal{V}|}$ naturally
implies 
$|\pi(T)-1| < 10^{-2}|\mathcal{V}|^{-1} \equiv \pi(T)> 1-\frac{1}{100 |\mathcal{V}|}$ and such principle can fit any situation in our framework.

So far, the nodes with difficulty level under the threshold of scheduler function $\pi(t)$ will be sampled and treated equally in model training stages. However, this could lead to a somewhat counterintuitive phenomenon:
those easy nodes (e.g., $Diff(v_i)<0.1$) are likely to be repeatedly used in model training, which deviates from the general learning principle of human. For example, college students who study calculus no longer need to review the simple addition and subtraction of numbers, but can still keep such capability when learning harder courses. From the view of optimization, we prefer at each iteration, the model can focus more on the difficult targets so as to better extend its border of capacity, while the repeatedly used easy samples are paid less attention.
To this end, we introduce a simple and dynamic weighting policy into our design, so that the model can further benefit from the learning of curriculum. Formally, at training step $t$, we define the weight of node $v_i$ as:
\begin{subnumcases}
{
	w_i^{(t)}=}
	1  \quad\quad\quad\quad \text{if} \quad Diff(v_i) \leq S_{t-1}\\
	1+\pi(t)  \quad \text{if} \quad S_{t-1}< Diff(v_i) \leq S_{t}\\
	0   \quad\quad\quad\quad \text{if} \quad Diff(v_i) > S_{t}\label{equ:zero}.
\end{subnumcases}
We do not consider weights for those nodes which haven't been included in training (i.e., $Diff(v_i) > S_{t}$) and set them as 0. And
our scheduler function $\pi(t)$ increases monotonically with its range in $[0,1]$. Therefore, at each training step, the loss of newly included nodes will be mildly emphasized with more weights, while others are applied normally with $w_i^{(t)}=1$. There are three things worth noting about the weighting design: 1) The weights of newly added nodes, which stand as the border of model's capacity at training step $t$, are positively correlated with the value of $\pi(t)$, since intuitively we consider, as $ \pi(t)$ becomes larger, the newly added nodes can be treated as a harder capacity bound for the model to breakthrough, and therefore requires larger weights. 2) When the curriculum is over, all the nodes will be assigned the same weights (i.e., 1) to recover normal training. 3) In addition to \eqnref{equ:dropout} which aims to prevent the unwanted feature fusion between easy and difficult nodes (i.e., $Diff(v_i) > S_{t}$) in forward propagation, \eqnref{equ:zero} seeks to block such noise from model in backward propagation.
Finally, we combine the weighting $w^{(t)} \in \mathbb{R}^{\mathcal{V}}$ with the objective function and formulate them as follows: 
\begin{align}\label{equ:loss}
\mathcal{L}(\theta)=\left\|w^{(t)} \left( \mathbf{X}^{(t)}_{(n+1):(n+\beta)}-\overline{\mathbf{X}}^{(t)}_{(n+1):(n+\beta)}  \right) \right\|_{2}^{2} ,
\end{align}
where $\mathbf{X}^{(t)}_{(n+1):(n+\beta)}$ and $\overline{\mathbf{X}}^{(t)}_{(n+1):(n+\beta)}$ indicate the ground truth and model prediction at training step $t$, and $\theta$ is the set of model parameters.

In our curriculum design, easier nodes are first fed to the model, helping it learn the fundamental ST relations while prevent it from being disturbed by the ``noisy'' data in early training stages, which is the most important time for training DNNs \cite{jastrzebski2020break}. After that, more complex samples are exposed to the model, allowing it to learn the higher-level abstraction in ST relations. The newly added nodes, which serve as a manner of regularization, both improve the generalization capability of the model
and allow the model to avoid over-fitting over the easy ST relations. As a result, at each training stage, the difficulty of samples
can stand at the border of the model's capacity, neither too easy nor too hard, so that the model can expand its border gradually. This is very similar to the idea of Continuation Methods \cite{allgower1980simplicial}, which suggests to first optimize a smoothed objective and gradually consider less
smoothing, with the intuition that a smooth version
of the problem can reveal the global picture.





 \section{Experiment}
 In this section, we evaluate our proposed STC-Dropout on a wide range of spatial-temporal datasets so as to demonstrate its general applicability: METR-LA \cite{li2017diffusion}, PEMSD7M \cite{yu2017spatio}, Covid-19 \cite{panagopoulos2021transfer} and NYC-Crime \cite{xia2021spatial}. Since this is the first attempt to introduce CL into ST modeling task, we validate the superiority of our strategy by comparing with the strong CL strategies from other fields: 1) \textbf{C-Dropout} (curriculum dropout) \cite{morerio2017curriculum} , 2) \textbf{C-Smooth} (curriculum by smoothing) \cite{sinha2020curriculum} and 3) \textbf{Anti-CL} \cite{shrivastava2016training} on state-of-the-art models : 1) \textbf{STGCN} \cite{yan2018spatial}, 2)
\textbf{Graph WaveNet} \cite{wu2019graph},  3) \textbf{ASTGCN} \cite{guo2019attention} and 4) \textbf{Z-GCNETs } \cite{chen2021z}. The four models we choose range from the classic ones to the latest ones, and they are all designed for ST tasks. \textit{Please refer to the appendix A.2-A.9 for more detailed descriptions of the datasets, baselines, experiment setup and extra results. }


\begin{table*}[h]
	\centering
	\caption{Performances of STC-Dropout on a wide range of spatial-temporal datasets.}\label{table:result_traffic_speed}
	\begin{adjustbox}{width=1\textwidth}
		\begin{tabular}{c|ccc|ccc}
			\toprule \multirow{2}{*}{ Model } & \multicolumn{3}{|c}{ METR-LA (15/30/60 min)}& \multicolumn{3}{|c}{ PeMSD7M (15/30/60 min)} \\
			\cline { 2 - 7 } & MAE & MAPE(\%) & RMSE  & MAE & MAPE(\%) & RMSE\\
			\hline 
			ST-GCN &2.89/3.50/4.62 &7.64/9.59/12.70 &5.75/7.28/9.40 &   2.12/2.72/3.35& 4.99/6.92/8.97& 4.09/5.53/6.62 \\
			+STC-Dropout &\bfseries2.67/3.28/4.37 &\bfseries7.37/9.22/12.22 &\bfseries5.54/6.83/9.06 &\bfseries2.03/2.61/3.27 &\bfseries4.83/6.85/8.64 &\bfseries3.85/5.30/6.38 \\
			\hline
			GW-Net & 2.70/3.05/3.53 &6.88/8.38/10.05 &5.17/6.24/7.37 &  2.08/2.66/3.23& 4.91/6.69/8.48& 3.94/5.31/6.35\\
			+STC-Dropout&\bfseries2.51/2.83/3.27 &\bfseries6.45/8.06/9.68 &\bfseries4.79/5.91/6.99 &\bfseries1.97/2.52/3.11 &\bfseries4.71/6.54/8.33 &\bfseries3.76/5.17/6.11 \\
			\hline
			ASTGCN &4.87/5.45/6.54 &9.24/10.14/11.62 &9.25/10.62/12.50 & 2.19/2.77/3.52& 5.42/7.31/9.18& 4.15/5.37/6.54\\
			+STC-Dropout&\bfseries4.70/5.26/6.40 &\bfseries8.93/9.74/11.32 &\bfseries9.02/10.12/12.26 
            &\bfseries2.02/2.68/3.40 &\bfseries5.30/7.25/8.90 &\bfseries4.06/5.15/6.20\\
			\hline
			Z-GCNETs & 2.62/2.99/3.32 & 6.79/8.45/9.97& 5.09/6.04/7.15 &1.91/2.35/2.98 & 4.86/6.52/8.18& 3.78/5.18/6.17\\
			+STC-Dropout &\bfseries2.50/2.78/3.05 &\bfseries6.48/8.03/9.56 &\bfseries4.79/5.69/6.82 &\bfseries1.85/2.21/2.84 &\bfseries4.61/6.22/8.01 &\bfseries3.50/4.81/5.98 \\
			\bottomrule
			\multirow{2}{*}{ Model } & \multicolumn{3}{|c}{ Covid-19 (3/5/7 days)}& \multicolumn{3}{|c}{ Crime (3/5/7 days)} \\
			\cline { 2 - 7 } & MAE & MAPE(\%) & RMSE  & MAE & MAPE(\%) & RMSE\\
			\hline 
			ST-GCN 	   &3.51/3.68/3.88 &2.29/2.36/2.53 &6.22/6.57/7.18 & 2.15/3.22/3.86 &1.40/2.10/2.52 &3.21/5.04/6.10 \\
			+STC-Dropout & \bfseries3.12/3.24/3.37 & \bfseries1.57/1.69/1.98 & \bfseries5.65/6.04/6.49 & \bfseries 2.07/3.14/3.80 & \bfseries1.31/2.05/2.40 & \bfseries3.15/4.97/5.98 \\
			\hline 	    
			GW-Net 	   &3.40/3.55/3.76 &2.26/2.30/2.37 &6.06/6.42/6.92 & 2.17/2.80/3.38 &1.42/1.83/2.21 &3.48/4.60/5.69 \\
			+STC-Dropout  & \bfseries2.89/3.18/3.26 & \bfseries1.49/1.65/1.81 & \bfseries5.56/5.81/6.24 & \bfseries 2.11/2.77/3.33 & \bfseries1.36/1.72/2.07 & \bfseries3.42/4.55/5.61 \\
			\hline      
			ASTGCN      &3.54/3.60/3.86 &2.31/2.33/2.52 &6.23/6.45/7.15 & 2.31/2.96/3.39 &1.78/1.99/2.28 &3.45/4.46/5.29 \\
			+STC-Dropout & \bfseries2.94/3.17/3.24 & \bfseries1.60/1.77/2.02 & \bfseries5.53/5.75/6.32 & \bfseries 2.24/2.91/3.32 & \bfseries1.65/1.91/2.17 & \bfseries3.35/4.37/5.21 \\
			\hline      
			Z-GCNETs    &3.27/3.42/3.59 &2.21/2.28/2.33 &5.95/6.22/6.74 &                            2.10/2.68/3.19 &1.31/1.85/2.12 &3.10/4.38/5.43 \\
			+STC-Dropout& \bfseries2.82/3.07/3.12 & \bfseries1.43/1.65/1.86 & \bfseries5.54/5.69/6.18 & \bfseries 2.06/2.61/3.12 & \bfseries1.22/1.78/2.04 & \bfseries3.08/4.29/5.30 \\
			
			\bottomrule
		\end{tabular}
		
	\end{adjustbox}
\end{table*}

		\begin{table}[h]
	\caption{Different curriculum learning strategies.}\label{tab:res2}
	\centering
	\begin{adjustbox}{width=0.39\textwidth}
		\setlength{\tabcolsep}{0.9mm}{\begin{tabular}{c|lccc}
				\hline Model & CL Strategy &  MAE &  MAPE(\%)   & RMSE \\
				\hline 			
				\multirow{4}{*}{{METR-LA}}& ST-GCN&  
				2.89&7.64&5.75\\
				&+Anti-CL & 3.04 &7.73 &5.91 \\
				&+C-Dropout&  2.75& 7.52& 5.69\\
				&+C-Smooth&  2.77& 7.60& 5.72\\
				&+STC-Dropout &\bfseries 2.67&\bfseries7.37&\bfseries5.54\\
				
				\hline
				\multirow{4}{*}{{PeMSD7M}}& GW-Net&  2.08&4.91&3.94\\
				&+Anti-CL & 2.16 &5.07 &4.11 \\
				&+C-Dropout&  2.03&4.91&3.92\\
				&+C-Smooth&  2.01&4.82&3.86\\
				&+STC-Dropout &\bfseries1.97&\bfseries 4.71&\bfseries3.76 \\
				\hline
				\multirow{4}{*}{{Covid-19}}& ASTGCN
				&3.54&2.31&6.23  \\
				&+Anti-CL& 3.55 &2.30 &6.23 \\
				&+C-Dropout& 3.23 &1.83 &6.06 \\
				&+C-Smooth&3.27 &1.78 &5.92  \\
				&+STC-Dropout &\bfseries 2.94&\bfseries 1.60&\bfseries 5.53\\
				\hline
				\multirow{4}{*}{{Crime}}& 
				Z-GCNETs&  2.10 &1.31 &3.10 \\
				&+Anti-CL &2.12 &1.34 &3.12 \\
				&+C-Dropout&  2.09 &1.25 &3.09 \\
				&+C-Smooth&2.07 &1.33 &3.09   \\
				&+STC-Dropout &\bfseries 2.06 &\bfseries1.22 &\bfseries3.08  \\
				\hline
		\end{tabular}}
	\end{adjustbox}
\end{table}

\subsection{Experimental Results}
We conduct extensive experiments on a wide range of ST datasets to demonstrate the superiority of STC-Dropout, which are listed in \tableref{table:result_traffic_speed}.
We can observe that generally,  SOTA models still have much room for improvement via the utilization of STC-Dropout.
We summarize the average improvements as follows:
\textit{MAE: 0.21 $\uparrow$, MAPE: 0.36 $\uparrow$, RMSE: 0.34 $\uparrow$} in METR-LA
, \textit{MAE: 0.47 $\uparrow$, MAPE: 0.63 $\uparrow$, RMSE: 0.61 $\uparrow$} in Covid-19,
\textit{MAE: 0.12 $\uparrow$, MAPE: 0.19 $\uparrow$, RMSE: 0.23 $\uparrow$} in PeMSD7M and \textit{MAE: 0.06 $\uparrow$, MAPE: 0.10 $\uparrow$, RMSE: 0.08 $\uparrow$} in Crime. Especially, according to recent traffic benchmarks \cite{li2021dynamic,jiang2021dl}, our performance gains on METR-LA and PeMSD7M can even match the performance improvements between two generations of models (e.g., DCRNN and GW-Net), and we achieve this by just using a simple plug-in to change the learning strategy (An intuitive reasoning in Appendix A.7).
\begin{figure}[t]
	\centering
	\includegraphics[width=1\linewidth]{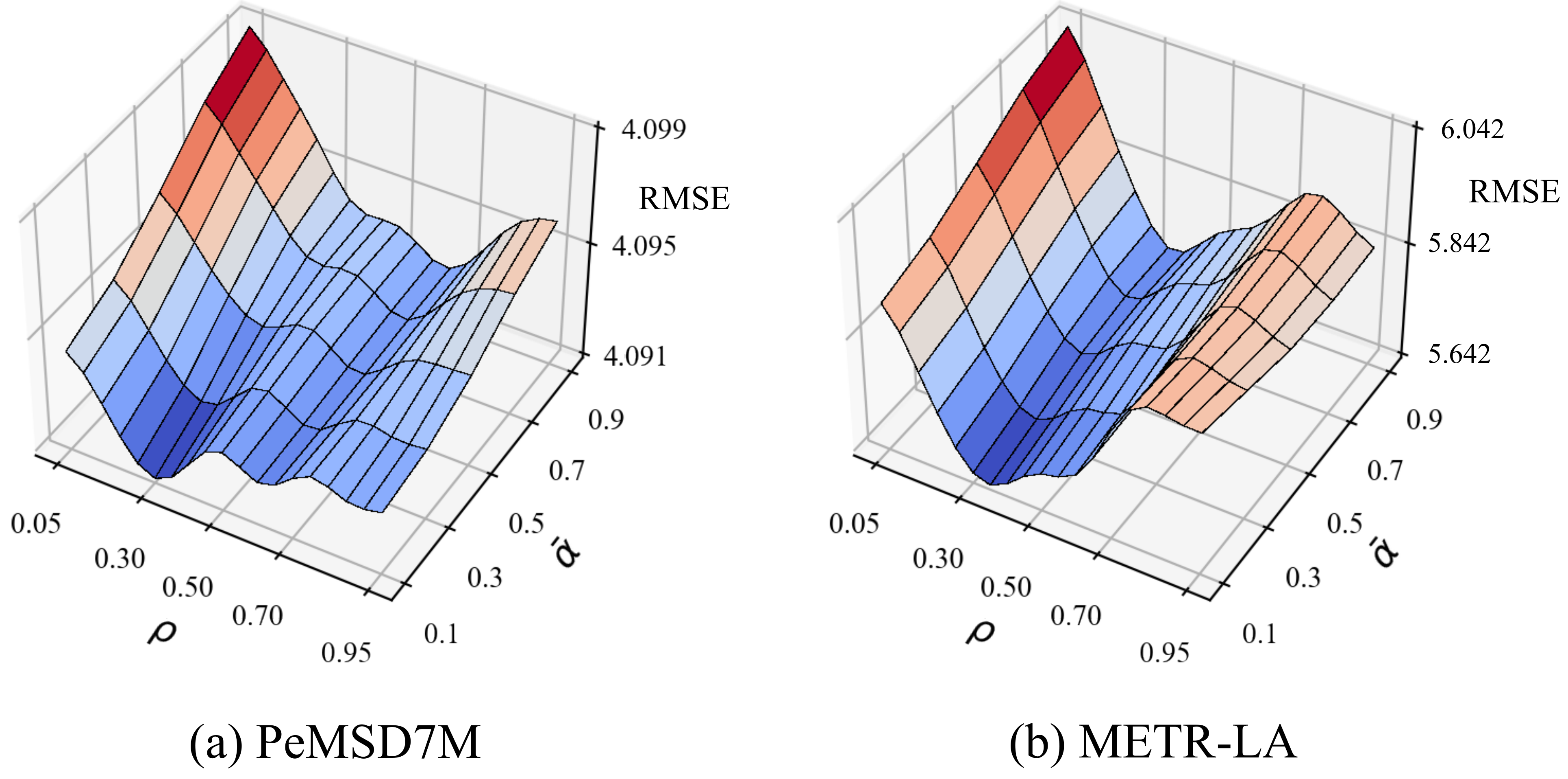}
	
	\caption{Grid search of hyper-paramerters on ST-GCN.}\label{fig:hypepramerter}
\end{figure}

We further horizontally compare STC-Dropout with several CL methods from other fields and the results are reported in \tableref{tab:res2} (and Appendix A.8). It's observed that both C-Dropout and C-Smooth enhance the baseline models by a certain margin, which again validates the idea of CL on ST modeling task. However, Anti-CL fails to outperform baselines. Therefore, we here discuss some reasons 
behind their performances and STC-Dropout's overwhelming improvements over others as follows:
\begin{itemize}[leftmargin=*]
	\item[$\bullet$] C-Dropout is a curriculum method proposed regarding the model itself, which progressively reduces the model's capacity in object classification tasks. Completely adopting such an idea into spatial-temporal modeling might be hard since the task is more complicated and requires a large network capacity to model high-level ST relations, especially at later training stages. The benefit of C-Dropout in experiment might come from the early stages of training, which offers better parameter initialization \cite{morerio2017curriculum}.
	
	\item[$\bullet$]  C-Smooth is first proposed in CNN using Gaussian kernel filters to ``blur'' the data and reduce noises. On the one hand, this might be too arbitrary without consideration of the ST relations. On the other hand, although we modify it to run on GNNs, this could arouse another problem called over-smoothing \cite{rong2019dropedge}. Conversely, STC-Dropout is specially developed for ST graph modeling and considers the ST relation at the stage of difficulty evaluation. 
 \item[$\bullet$] Anti-CL is another popular data selection strategy opposite to CL (i.e. hard sample first). A well-accepted insight \cite{chang2017active} about whether to choose CL or Anti-CL points out that CL is more preferable for noisy datasets to improve model robustness and Anti-CL is more beneficial for cleaner datasets to accelerate convergence. In light of this, we attribute the inferior performances of Anti-CL to the chaotic nature of ST data.  
\end{itemize}

\subsection{Ablation Study}
\noindent\textbf{Comparison with Heuristic Curriculum Strategy.}
To validate the effectiveness of STC-Dropout, we compare it with multiple heuristic curriculum designs (without ST correlation assumptions) to forecast future 15 min's traffic speed using ST-GCN. We perform the curriculum both in the high-level hidden space (as STC-Dropout does) and directly at the input (corresponding to the discussion in the first paragraph of \textbf{Difficulty Evaluation}). Two types of heuristic curriculum schemes are used here: one defines the difficulty of nodes via model predicting loss \cite{guo2018curriculumnet} and the other seeks the complex samples through outlier detection \cite{zhang2019decomposition}. For predicting loss we apply a basic time series prediction models: \textbf{LR} (Linear Regression) as the difficulty indicator, while for outlier detection, we use the popular corresponding algorithms: \textbf{LoF} (Local Outlier Factor)  \cite{breunig2000lof} and \textbf{IF} (Isolation Forest) \cite{liu2008isolation}. These above-mentioned methods will rank the difficulty of nodes either in the hidden dimension or at the input space, and then they follow the same schema of STC-Dropout to dynamically dropout difficult nodes at early stages. We report the results in \tableref{tab:her}, where we can discern that STC-Dropout, which we designed with ST relation assumptions, yields the best outcome, and the curriculum performed in hidden states is generally better than those in the input space.

Further, we consider several variants of STC-Dropout including the origin \textbf{Dropout} without curriculum \cite{hinton2012improving}, \textbf{STC-Dropout-Mean}: set the dropout values to its neighbor's mean instead of zeros, \textbf{SC-Dropout}: only utilizes spatial difficulty measurer (Formula \ref{equ:spatial}), and  \textbf{TC-Dropout}: only uses temporal difficulty measurer (Formula \ref{equ:temporal}). The weighting design is also included in the ablation. The comparison results are shown on \tableref{table:ablation}.   
Regarding the variants, there is no improvement over the standard STC-Dropout on ST-GCN. 
\begin{table}[t]
	\centering
	\caption{Comparison with heuristic curriculum strategy}\label{tab:her}
	\begin{adjustbox}{width=0.35\textwidth}
		\setlength{\tabcolsep}{0.9mm}{\begin{tabular}{l|l l c  c c c }
				\hline 	\multirow{2}{*}{Level} & \multirow{2}{*}{Models}   & \multicolumn{3}{c}{METR-LA}  \\
				
				&\cline{2-4}   &  & {\small MAE } & {\small MAPE(\%)} & {\small RMSE} \\
				\hline\multirow{4}{*}{\rotatebox[origin=c]{90}{Hidden }}
				&ST-GCN			 &2.89&7.64&5.75\\
				&+C-LoF			 &2.73& 7.48&5.66\\
				&+C-IF			 &2.78& 7.58&5.71\\
				&+STC-Dropout		 &\bfseries 2.67&\bfseries7.37&\bfseries5.54 \\
				\hline	\multirow{4}{*}{\rotatebox[origin=c]{90}{Input }}
				&+C-LR			 &2.88& 7.67&5.77\\
				&+C-LoF			 &2.76& 7.53&5.70\\ 
				&+C-IF			 &2.84& 7.63&5.74\\
				&+STC-Dropout	&\bfseries2.69&\bfseries7.42&\bfseries 5.61\\
				\hline
		\end{tabular}}

		\label{table:ablation}
	\end{adjustbox}
\end{table}


\begin{table}[h]
	\centering
	\caption{Ablation Study} 
	\begin{adjustbox}{width=0.4\textwidth}
		\setlength{\tabcolsep}{1.4mm}{\begin{tabular}{l l c c c c }
				\hline 	 \multirow{2}{*}{Models}   & \multicolumn{3}{c}{METR-LA}  \\
				
				\cline{2-4}   &   {\small MAE } & {\small MAPE(\%)} & {\small RMSE} \\
				\hline
				ST-GCN				 &2.89&7.64&5.75\\
				+Dropout			&2.87& 7.63&5.74 \\
				+STC-Dropout-Mean		&2.79& 7.60&5.71 \\
				+SC-Dropout	 &2.76& 7.51&5.66\\
				+TC-Dropout	 &2.73& 7.49&5.68\\
				+w/o. weight 		 &2.70& 7.44&5.58\\
				+STC-Dropout			  &\bfseries 2.67&\bfseries7.37&\bfseries5.54 \\
				\hline
		\end{tabular}}
		
	\end{adjustbox}
	\label{table:ablation}
\end{table}

\noindent\textbf{Hyper-parameter Analysis.}
Last but not least, we explore the hyper-parameters sensitivity to $\rho$, which controls the radius of the $d^{(l)}$-dimensional ball, and $\bar{\alpha}$, which specifies the initial retain rate of nodes.  The results of hyper-parameter search from $\rho \times \bar{\alpha}  \in [0.05,0.10,\cdots,0.95]  \times  [0.1,0.3, \cdots 0.9]$ on ST-GCN using both PeMSD7M and METR-LA datasets are depicted in  \figref{fig:hypepramerter}. As we can see, the model performance is negatively related to the value of $\bar{\alpha}$. Thus, it is recommended to set its value to be relatively small to ensure only the easy samples can be retained in the early model training. Concerning $\rho$, we encourage maintaining $\rho \in [0.25,0.4]$ for desirable performances. 

\begin{figure}[h]
	\centering
	\includegraphics[width=0.8\linewidth]{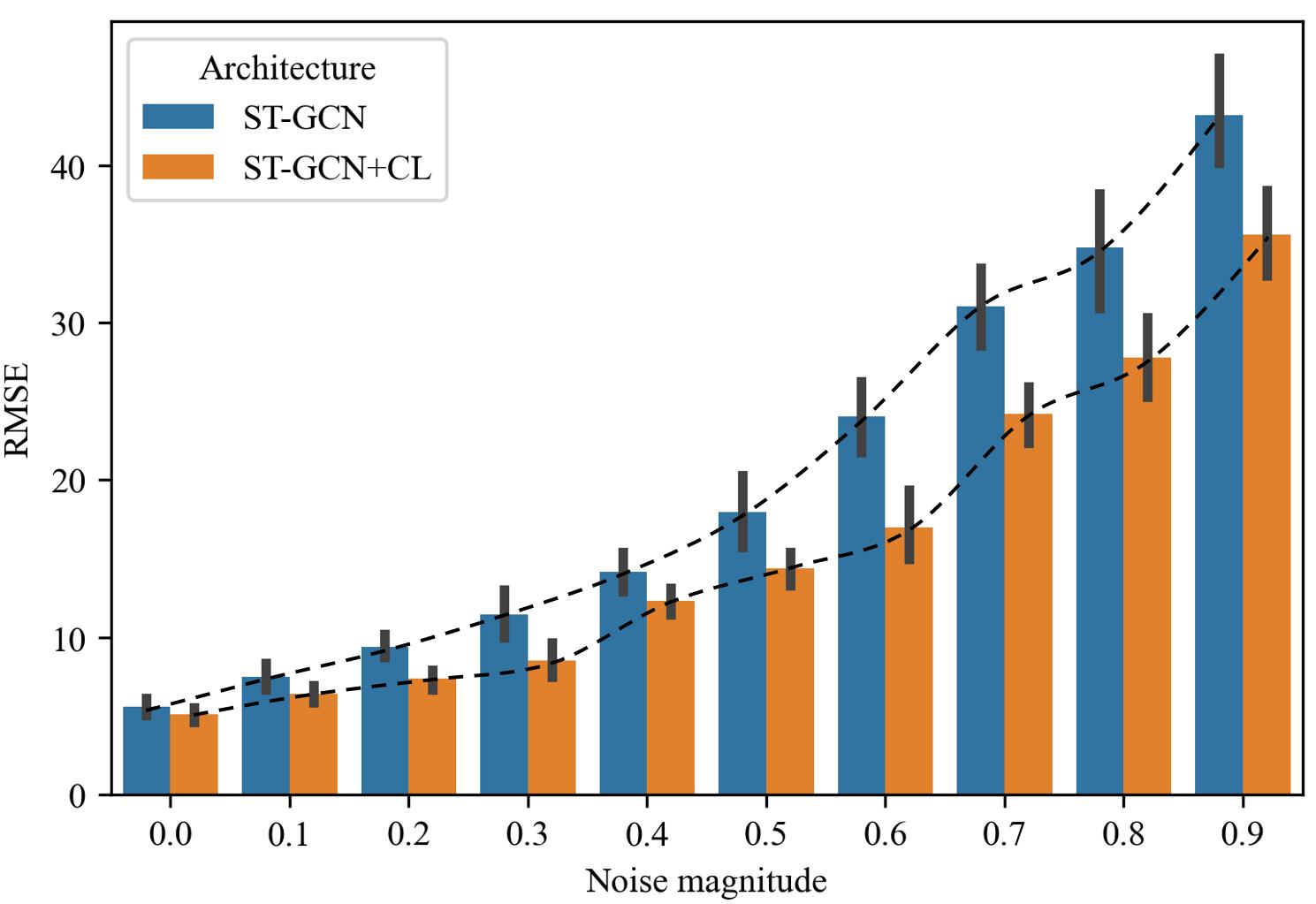}
	
	\caption{RMSE of ST-GCN with different noise magnitudes imposed on METR-LA.(More in Appendix A.9) }
	\label{fig:noise}
\end{figure}

\noindent\textbf{When Does CL Take The Best Effect.}\label{sec:robustness}
We observe an interesting phenomenon in \tableref{table:result_traffic_speed}: our method brings more improvements to METR-LA and Covid-19 than to the others. This is
somewhat reasonable since CL mainly works by excluding the negative impacts of difficult samples to the model in the early training. So its effect will decrease as the dataset becomes easy (e.g., for ST-GCN, METR-LA is easier than Crime since it has lower prediction error). We further validate our assumption via an experiment where we disturb the training data 
as follows:
$$
\hat{z}=(1+\delta)z, \quad \delta \sim uniform(\bigtriangleup, \bigtriangleup).
$$
The noise magnitude is controlled by $\bigtriangleup$ and \figref{fig:noise} shows our result. We can see that the as the noise magnitude becomes larger, our method gradually brings more improvements. This also suggests that our method can enhance robustness for models when training with noisy samples.


\section{CONCLUSION}
This paper designs a novel curriculum strategy called ST-Curriculum Dropout, especially for spatial-temporal graph modeling. By defining the complex score of nodes based on their $G$-neighbor consistency (representing the difficulty of spatial relations) and data distribution density (representing the difficulty of temporal patterns), we gradually expose the complex nodes to the network to form a curriculum. Experimental results show the effectiveness of our design. One potential problem of our method could be the computation of pairwise distance. Although the cost is acceptable, this can be further alleviated by using Faiss\footnote{https://github.com/facebookresearch/faiss} to accelerate computing. In future work, we would like to extend the STC-Dropout into further applications such as multi-variate time series.

\section{Acknowledgment}

We are grateful to anonymous reviewers for their helpful comments. We
additionally would like to thank Lei Huang for his help. This work was partially  supported by the grants of National Key Research and Development Program of China (No. 2018AAA0101100),   National Key Research and Development Project (2021YFB1714400) of China and  Guangdong Provincial Key Laboratory (2020B121201001), and Grant-in-Aid for Scientific Research B (22H03573) of Japan Society for the Promotion of Science (JSPS).

{\small

	\bibliographystyle{aaai23}
	\bibliography{citation}
}
\clearpage
\appendix 
\setcounter{figure}{0}    
\setcounter{table}{0}
\setcounter{equation}{0}

\section{Appendix}\label{appendix}

\subsection{A.1 Related Work}
\noindent\textbf{Curriculum Learning.}
The basic idea of curriculum learning is to let the models first learn from easy samples and work their way up to complex samples and knowledge \cite{bengio2009curriculum}. A well-written survey \cite{wang2021survey} classifies the taxonomy of curriculum learning into data-level and criteria-level. The data-level approach follows the natural definition of curriculum learning through sampling \cite{jiang2014easy}, weighting \cite{kumar2010self} and specific iterative \cite{spitkovsky2009baby} to select the most suitable examples for each training step. The other bunch of study considers arranging the curriculum according to some criteria which can be any design element in training a machine learning model. A novel training strategy \cite{karras2017progressive} has been designed to gradually increase the model's capacity by adding new neural units. Reference \cite{sinha2020curriculum} combines the idea of adversarial training with curriculum learning by using Gaussian kernel to smooth the noise in CNN feature maps at early stages of training. Reference \cite{morerio2017curriculum} introduces the dynamically stochastic dropout in curriculum learning by moderately increasing the dropout rate during training to handle co-adaptions in DNN. 


\noindent\textbf{Spatial-temporal Graph Modeling.}
The task of spatial-temporal graph modeling intends to forecast the future single/multi-step signals according to the historical observation and spatial graph information. Lots of applications including taxi demand prediction \cite{yang2021dual}, human action recognition \cite{peng2020learning}, traffic speed forecast \cite{wu2020connecting} and driver maneuver anticipation \cite{jain2016structural} can be categorized to this case. The 
primary challenge of spatial-temporal graph modeling lies in simultaneously capturing spatial and temporal dependencies from dynamic node-level inputs and a fixed graph structure. Recent studies to solve this problem mainly integrate the graph convolution
networks (GCN) into existing models \cite{yan2018spatial,li2017diffusion,seo2018structured,yu2017spatio}. In \cite{seo2018structured}, the graph convolution has been utilized to filter the node-level inputs and hidden state before passing them to a recurrent unit. A diffusion graph convolution method is further devised \cite{li2017diffusion} to better capture the spatial dependency. Reference \cite{yu2017spatio,yan2018spatial} combine GCN with standard 1D convolution layers instead of recurrent units to overcome the drawbacks of RNN-based approaches that tend to experience gradients explosion in long sequences scenario. The attention mechanisms are also explored by using a convolution
subnetwork to control the importance of features from GCN \cite{zhang2018gaan} or directly employing graph attention \cite{guo2019attention}. 
More recently, researchers further propose a variety of work to address the common problem of  location-characterized (local) patterns \cite{bai2020adaptive,lin2021conditional}. 
While all the abovementioned efforts focus on building more complex models to enhance their abilities to capture spatial-temporal dependencies, our contribution to this task lies in a different direction. Instead of proposing yet another ST model, we explore the possibility of simply using a novel CL training strategy to let the model learn from more meaningful and discriminative features and finally result in better overall performances. To the best of our knowledge, we are the first work to explore such ideas.

\subsection{A.2 Datasets}

The METR-LA dataset collects four months of traffic speed data using 207 sensors on the highways of Los Angeles. PEMSD7M collects two months statistics of traffic speed using 228 sensors in Caltrans Performance Measurement System (PeMS). These two datasets contain time-stamped characteristics of the average speed and the sensors' geographic locations, where they collect the speed information.

The COVID-19 dataset used by \cite{panagopoulos2021transfer} is first released by Facebook in the scope of Data
for Good program\footnote{https://dataforgood.facebook.com/dfg/tools}, which intends to provide opportunities for researchers to investigate the dynamics of COVID-19, we here focus on the data of an European country: France. In this dataset, each node indicates a specific  region of a country, and the weight of the edges represents the corresponding volume of human flow from one region to another. Since the existing ST models usually set edge weights to be static, we here utilize the average human mobility between regions as their weights. 

The Crime dataset is collected from New York City's OpenData\footnote{https://data.cityofnewyork.us/Public-Safety/NYC-crime/qb7u-rbmr} that includes crimes reported to the New York City Police Department (NYPD). In this dataset, we divide the NYC into 256 disjoint spatial regions and formulated them as a graph with each node indicating a $3km \times 3km$ spatial grid. 

The statistics of the datasets are reported in \tableref{tab:dataset}. We follow the data pre-processing conducted in \cite{li2017diffusion} to use a Gaussian kernel as follows:
$\hat{w}_{i j}=\exp \left(-\left(\frac{w_{i j}}{\sigma}\right)^{2}\right)$ to smooth the weights of edges and build the adjacency matrix. The inputs are the distances between and $\sigma$ is a normalizing factor. In terms of the graph signals, we use the $z$-score normalization. 

\begin{table*}[h]
	\centering
	\caption{ Dataset statistics.}
	\begin{tabular}{l|cccc}
		\hline Tasks & METR-LA & PeMSD7M& Covid-19& Crime\\
		\hline Start time & $10 / 1 / 2014$ & $5 / 1 / 2012$ & 2/20/2020 &1/1/2014 \\
		End time & $12 / 31 / 2014$ & $6 / 30 / 2012$ &20/6/2020 & 12/31/2015\\
		Sample rate & 5 minutes & 1 minutes &1 day & 1 day\\
		Timesteps & 27403 & 12672 &123&730 \\
		Sensors/Regions & 207 & 228  &55&256\\
		\hline Training size & 16783 & 8870&94& 510 \\
		Validation size &  2397 &  1267 &12 & 73 \\
		Testing size & 4795 & 2549 &24 & 146\\
		Output length & 3, 6, 12 & 3, 6, 12 & 3, 5, 7 & 3, 5, 7 \\
		\hline
	\end{tabular}\label{tab:dataset}
\end{table*}

\subsection{A.3 Baselines Description and Setups}

\begin{itemize}[leftmargin=*]
	\item[$\bullet$] \textbf{Curriculum Dropout} \cite{morerio2017curriculum}: Curriculum Dropout gradually increases the dropout rate to reduce the model capacity to form an easy-to-hard curriculum for the model. We use the recommended hyper-parameters: $\theta(0)=1$ and $\gamma= 10/T$  in the curriculum dropout as the author suggested.
	\item[$\bullet$] \textbf{Curriculum by Smoothing} \cite{sinha2020curriculum}: The strategy implements the curriculum with a Gaussian smoothing filter to reduce the ``noise'' of data in early training stages on CNNs, and gradually anneals the value of $\sigma$ in Gaussian filter to recover normal training. We extend such settings to GCNs. The initial value of $\sigma^{(0)}$ is set to 1, and gradually decreases every 10 epochs.
	\item[$\bullet$]
	\textbf{Anti-CL} \cite{shrivastava2016training}
	: Anti-Curriculum is a strategy opposite to Curriculum Learning. In our experiments, we simply reverse the scheduler function to simulate Anti-CL.
	\item[$\bullet$] \textbf{ST-GCN} \cite{yan2018spatial}: ST-GCN integrates the  graph convolution and 1D convolution. It contains two ST-Conv blocks with (12, 16, 64) and (64, 16, 64) respectively. 
	
	\item[$\bullet$] \textbf{Graph WaveNet} \cite{wu2019graph}: GW-Net combines graph convolution with dilated causual convolution. In Graph WaveNet, the number of layers and blocks are defaulted as 4 and 2.

	\item[$\bullet$] \textbf{ASTGCN} \cite{guo2019attention}:  ASTGCN fuses the spatial-temporal attention mechanism and the graph convolution together. Each graph convolution filter of ASTGCN has 64 kernels.

 	\item[$\bullet$] \textbf{Z-GCNETs } \cite{chen2021z}: Z-GCNETs uses the most salient time-conditioned topological information of the data and formulates zigzag persistence into a time-aware graph
     convolutional networks (GCNs). The model structure of Z-GCNETs follows the original paper.
 \end{itemize}

\subsection{A.4 Experimental Setups}
All baselines in this paper are implemented based on the Pytorch framework with   eight GeForce GTX 2080Ti graphics cards and one Intel(R) Xeon(R) Silver 4216 CPU @ 2.10GHz. The datasets are divided into three parts in chronological order for training, validation, and testing with the ratio of $7:1:2$. 
An Adam optimizer with initialized learning rate $10^{-3}$ is applied to all models in experiments and
the batch size was set to 64 by default. The training algorithm would
either be early-stopped if the validation error converges within 10 epochs or be stopped after 200 epochs. In order to verify the performance of the STC-Dropout, the evaluation indicators selected in this paper are Root Mean Squared Error (RMSE), Mean Absolute Error (MAE), and Mean Absolute Percentage Error (MAPE). The random seed is predefined for the consistency of the model's initial parameters, based on which the reproducibility and the reliability of the experimental results can be guaranteed. Moreover, we present the prediction error of Covid-19 and Crime datasets as the accumulated error for 3/5/7 days so as to better illustrate the model performances.

\begin{figure}[t]
	\centering
	\includegraphics[width=1\linewidth]{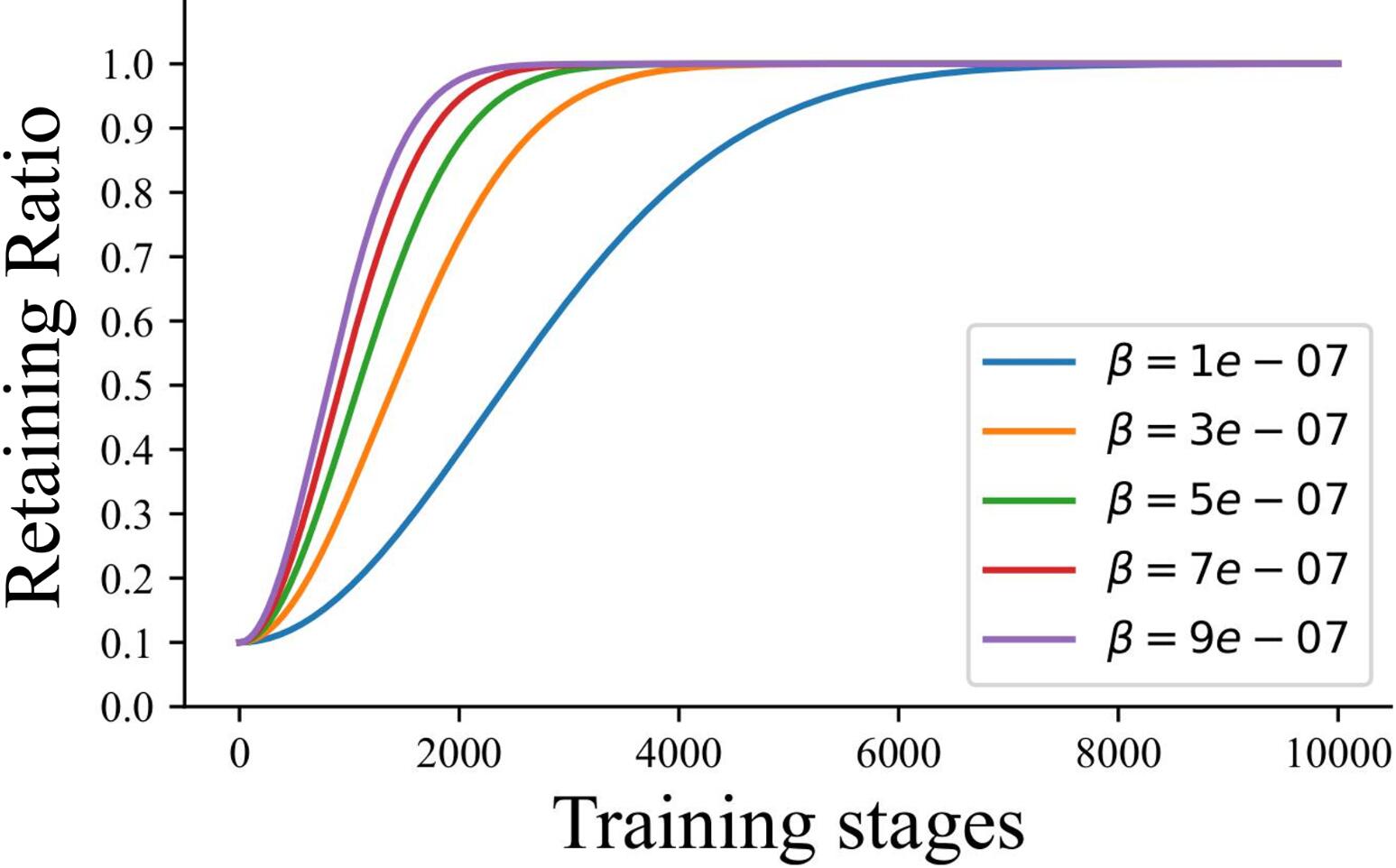}
	\caption{ visualizing the curriculum scheduler function $\pi$ with different settings. 
	 }\label{fig:curriculum_function}
\end{figure}

\subsection{A.5 
Curriculum Scheduler Function}

We further visualize the curriculum scheduler function $\pi$ defined in \eqnref{eqn:curriculum_function}. We can observe from \figref{fig:curriculum_function} that the curve tends to be smoother with the increase of $\beta$. Recall that $\beta$ is heuristically set as $\frac{10^{3}}{T\times|\mathcal{V}|}$, so as to satisfy $|\pi(T)-1| < 10^{-2}|\mathcal{V}|^{-1}$. Take METR-LA for an example, assuming $|\mathcal{V}|=207$, and $T=10000$, $\beta$ can be defined as $\frac{1}{2070}$. At each training stage $t$, the scheduler function $\pi$ will give a ratio $\pi(t) \in (0,1)$ to retain the easiest $\pi(t)\%$ of nodes for training.

\subsection{A.6  Evaluation Metric}
 RMSE (Root Mean Square Error), MAE (Mean Absolute Error), and MAPE (Mean Absolute Percentage Error) are used in this paper to evaluate the performance of baselines:

\begin{align*}
\left\{\begin{aligned}
R M S E &=\sqrt{\frac{1}{\mathcal{T}_{\text {test }}} \sum_{t}^{\mathcal{T}_{\text {test }}}\left\|\hat{\mathbf{X}}_{(t+1):(t+\beta)}-\mathbf{X}_{(t+1):(t+\beta)}\right\|^{2}} \\
M A E &=\frac{1}{\mathcal{T}_{\text {test }}} \sum_{t}^{\mathcal{T}_{\text {test }}}\left|\hat{\mathbf{X}}_{(t+1):(t+\beta)}-\mathbf{X}_{(t+1):(t+\beta)}\right| \\
M A P E &=\frac{1}{\mathcal{T}_{\text {test }}} \sum_{t}^{\mathcal{T}_{\text {test }}}\left|\frac{\hat{\mathbf{X}}_{(t+1):(t+\beta)}-\mathbf{X}_{(t+1):(t+\beta)}}{\mathbf{X}_{(t+1):(t+\beta)}+\varepsilon}\right|
\end{aligned}\right.
\end{align*}
where $\mathcal{T}_{\text {test }}$ is the test datasets, $\hat{\mathbf{X}}_{t}$ and $\mathbf{X}_{t}$ are the predicted and ground truth tensor respectively, and $\varepsilon$ is a constant to prevent dividing by zero.

\subsection{A.7 STC-Dropout's Relation with Heterophily Graph}
Essentially, the graph with corrupted spatial-temporal relations can be viewed as a special form of heterophily graph, where the connected nodes belong to different classes (under ST assumptions, the signals of connected nodes don't share similarity). In the literature of heterophily graphs, a large amount of models are proposed to solve this problem, such as GAT \cite{velivckovic2017graph}
which aims to use attention to study the nodes of the same class more directly, and H2GCN
\cite{zhu2020beyond}, which changes its aggregation function to capture long-range correlations. Correspondingly, in ST mining work, various work like ASTGCN \cite{guo2019attention} , GMAN \cite{zheng2020gman} and AGCRN \cite{bai2020adaptive} are proposed. Fundamentally, these work all try to design new learning paradigm for GCN models (e.g., the model does not aggregate information from traditional neighbors but to use attention to decide which node gives the most informative information). Our STC-Dropout can also be viewed as a new learning paradigm for GCN models. In fact, if we integrate STC-Dropout into some GCNs, we can actually propose a new model for ST graph modeling. However, we consider it unnecessary because such design will be less flexible and hard to extend. Therefore, we instead use a simple plug-in to achieve the same target and it can be applied to any canonical ST graph models.

\begin{table}[h]
	\caption{Different curriculum learning strategies.}\label{apx:tab1}
	\centering
	
	\setlength{\tabcolsep}{1.4mm}{\begin{tabular}{c|lccc}
			\hline Model & CL Strategy &  MAE &  MAPE(\%)   & RMSE \\
			\hline 			
			\multirow{4}{*}{{METR-LA}}& 
			GW-Net&  	2.70& 6.88& 5.17 \\
			&+Anti-CL& 2.89 &7.02 &5.31 \\
			&+C-Dropout& 2.60& 6.62& 4.91 \\
			&+C-Smooth& 2.64& 6.72& 4.99 \\
			&+STC-Dropout &\bfseries2.51&\bfseries 6.45& \bfseries4.79 \\
			\hline
			\multirow{4}{*}{{PeMSD7M}}
			& ASTGCN&  2.19& 5.42& 4.15  \\
			&+Anti-CL& 2.26 &5.64 &4.26\\
			&+C-Dropout&  2.12 & 5.35 & 4.12  \\
			&+C-Smooth&2.10& 5.31& 4.10  \\
			&+STC-Dropout& \bfseries 2.02& \bfseries5.30& \bfseries4.06 \\
			\hline
			\multirow{4}{*}{{Covid-19}}& Z-GCNETs&3.27&2.21&5.95  \\
			&+Anti-CL& 3.30 &2.25 &6.01 \\
			&+C-Dropout& 2.69 &1.88 &5.62  \\
			&+C-Smooth&2.74 &2.03 &5.78   \\
			&+STC-Dropout &\bfseries 2.28& \bfseries1.43& \bfseries5.54\\
			\hline
			\multirow{4}{*}{{Crime}}
			& ST-GCN &  2.15&1.40&3.21	\\
			&+Anti-CL&2.18 &1.48 &3.23  \\
			&+C-Dropout&   2.13 &1.38 &3.20 \\
			&+C-Smooth&  2.14 &1.35 &3.17  \\
			&+STC-Dropout & \bfseries 2.07& \bfseries 1.31& \bfseries3.15  \\
			\hline
	\end{tabular}}
	
\end{table}

\subsection{A.8 Comparison with Different Curriculum Learning Strategies.}
\tableref{apx:tab1} shows the extra results of different curriculum learning strategies.

\subsection{A.9 When Does CL Take The Best Effect.}
\begin{figure}[h]
	\centering
	\includegraphics[width=1\linewidth]{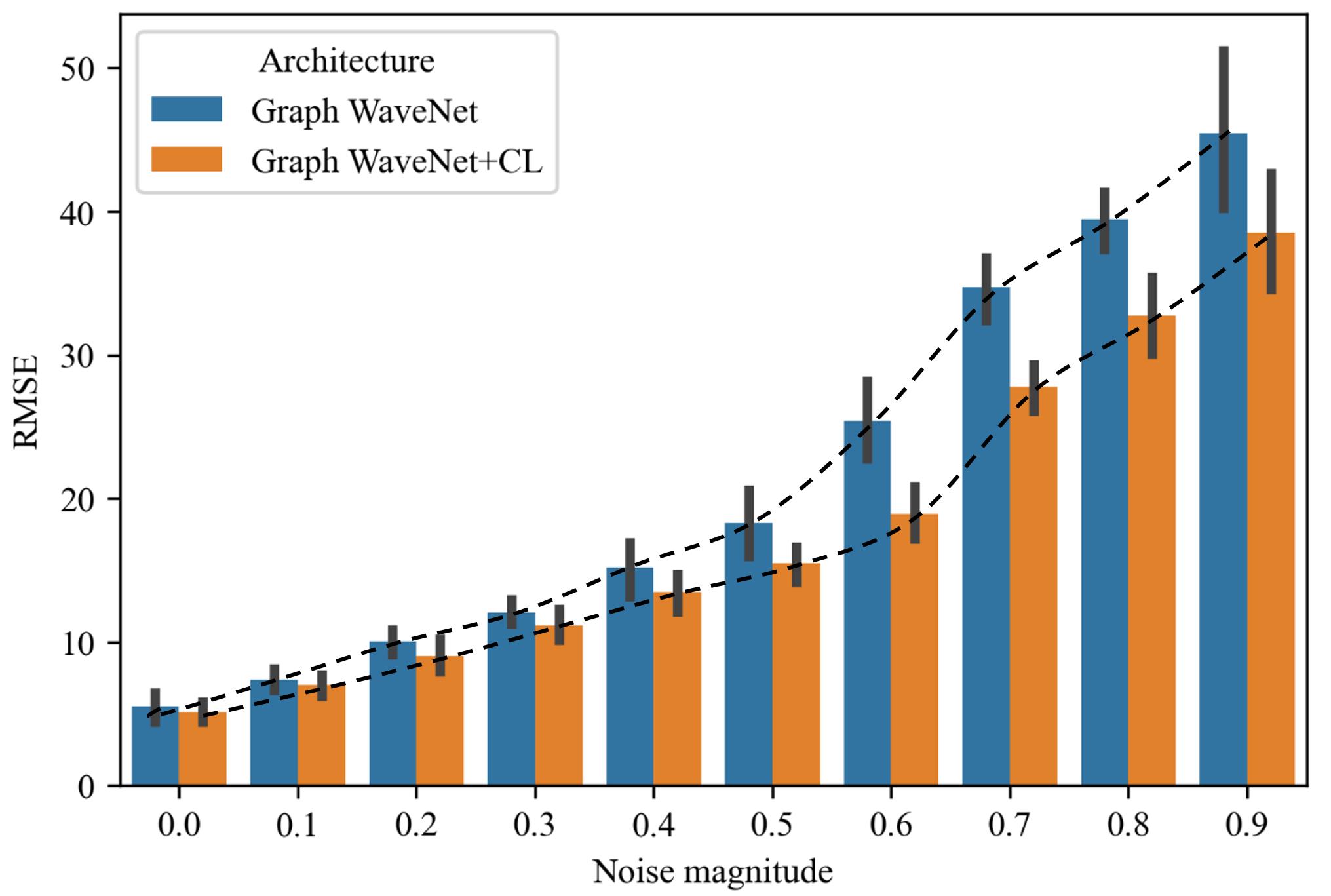}
	
	\caption{  RMSE of Graph WaveNet with different noise magnitudes imposed on the METR-LA.}
	\label{fig:noise2}
\end{figure}

\figref{fig:noise2} shows the performance of Graph WaveNet in METR-LA with different noise magnitudes. 
\end{document}